\documentclass[11pt,a4paper]{article}
\usepackage[hyperref]{acl2020}
\usepackage{times}
\usepackage{latexsym}

\usepackage[hang,flushmargin]{footmisc}  
\usepackage{amsfonts,amsmath,amssymb}
\usepackage[T1]{fontenc}  
\usepackage{bold-extra}   
\usepackage{microtype}    
\usepackage{subcaption}
\usepackage{graphicx}
\usepackage{multirow}
\usepackage{enumitem}
\usepackage{url}

\aclfinalcopy 

\newcommand\LN{\linebreak\noindent}

\title{Transformer-based Context-aware Sarcasm Detection in\\Conversation Threads from Social Media}

\author{Xiangjue Dong \\
  Computer Science \\
  Emory University \\
  Altanta, GA, USA \\
  \texttt{\small xiangjue.dong@emory.edu} \\\And
  Changmao Li \\
  Computer Science \\
  Emory University \\
  Altanta, GA, USA \\
  \texttt{\small changmao.li@emory.edu} \\\And
  Jinho D. Choi \\
  Computer Science \\
  Emory University \\
  Altanta, GA, USA \\
  \texttt{\small jinho.choi@emory.edu} \\}

\date{}

\begin{document}
\maketitle

\begin{abstract}

We present a transformer-based sarcasm detection model that accounts for the context from the entire conversation thread for more robust predictions.
Our model uses deep transformer layers to perform multi-head attentions among the target utterance and the relevant context in\LN the thread.
The context-aware models are evaluated on two datasets from social media, Twitter and Reddit, and show 3.1\% and 7.0\% improvements over their baselines.
Our best models give the F1-scores of 79.0\% and 75.0\% for the Twitter and Reddit datasets respectively, becoming one of the highest performing systems among 36 participants in this shared task.

\end{abstract}
\section{Introduction}
\label{sec:introduction}

Sarcasm is a form of figurative language that implies a negative sentiment while displaying a positive sentiment on the surface \cite{joshi-etal:2017}. 
Because of its conflicting nature and subtlety in language, sarcasm detection has been considered one of the most challenging tasks in natural language processing.
Furthermore, when sarcasm is used in social media platforms such as Twitter or Reddit to express users' nuanced intents, the language is often full of spelling errors, acronyms, slangs, emojis, and special characters, which adds another level of difficulty in this task.


Despite of its challenges, sarcasm detection has recently gained substantial attention because it can bring the last gist to deep contextual understanding for various applications such as author profiling, harassment detection, and irony detection \cite{hee-etal:2018}. 
Many computational approaches have been proposed to detect sarcasm in conversations \cite{ghosh-etal:2015,joshi-etal:2015,joshi-etal:2016}.
However, most of the previous studies use the utterances in isolation, which makes it hard even for human to detect sarcasm without the contexts.
Thus, it's essential to interpret the target utterances along with contextual information comprising textual features from the conversation thread, metadata about the conversation from external sources, or visual context \cite{bamman-smith:2015,ghosh-etal:2017,ghosh-veale:2017,ghosh-etal:2018}. 



This paper presents a transformer-based sarcasm detection model that takes both the target utterance and its context and predicts if the target utterance involves sarcasm.
Our model uses a transformer encoder to coherently generate the embedding representation for the target utterance and the context by performing multi-head attentions (Section~\ref{sec:approach}).
This approach is evaluated on two types of datasets collected from Twitter and Reddit (Section~\ref{sec:task-description}), and depicts significant improvement over the baseline using only the target utterance as input (Section~\ref{sec:experiments}).
Our error analysis illustrates that the context-aware model can catch subtle nuance that cannot be captured by the target-oriented model (Section~\ref{sec:analysis}).

%
\section{Related Work}
\label{sec:related-work}

Just as most other types of figurative languages are, sarcasm is not necessarily complicated to express but requires comprehensive understanding in context as well as commonsense knowledge rather than\LN its literal sense~\cite{hee-etal:2018}.
Various approaches have been presented for this task.


Most earlier works had taken the target utterance without context as input.
Both explicit and implicit incongruity features were explored in these works \cite{joshi-etal:2015}. 
To detect whether certain words in the target utterance involve sarcasm, several approaches based on distributional semantics were proposed \cite{ghosh-etal:2015}. 
Additionally, word embedding-based features like distance-weighted similarities were also adapted to capture the subtle forms of context incongruity \cite{joshi-etal:2016}. 
Nonetheless, it is difficult to detect sarcasm by considering only the target utterances in isolation.


\noindent Non-textual features such as the properties of the author, audience and environment were also taken into account \cite{bamman-smith:2015}. 
Both the\LN linguistic and context features were used to distinguish between information-seeking and rhetorical questions in forums and tweets \cite{oraby-etal:2017}. 
Traditional machine learning methods such as Support Vector Machines were used to model sarcasm detection as a sequential classification task over the target utterance and its surrounding utterances \cite{Wang-etal:2015}. 
Recently, deep learning methods using LSTM were introduced, considering the prior turns \cite{ghosh-etal:2017} as well as the succeeding turns \cite{ghosh-etal:2018}. 



\section{Data Description}
\label{sec:task-description}

Given a conversation thread, either from Twitter or Reddit, a target utterance is the turn to be predicted,\LN whether or not it involves sarcasm, and the context is an ordered list of other utterances in the thread.
Table~\ref{tab:sarcasm-examples} shows the examples of conversation threads where the target utterances involve sarcasm.\footnote{Note that the target utterance can appear at any position of the context although its exact position is not provided in this year's shared task data.}

\begin{table}[htbp!]
\centering

\begin{subtable}{\columnwidth}
\centering\resizebox{\columnwidth}{!}{
\begin{tabular}{c||l}
 & \multicolumn{1}{c}{\bf Utterance} \\ 
\hline\hline
\multirow{2}{*}{\texttt{C$_1$}}
  & This feels apt this morning but I don't feel fine ... \\
  & <URL> \\
\hline
\multirow{2}{*}{\texttt{C$_2$}}
 & @USER it is what's going round in the heads of \\
 & many I know ...\\
\hline
\multirow{3}{*}{\texttt{T}}
 & @USER @USER I remember a few months back\\
 & we were saying the Americans shouldn't tell us\\
 & how to vote on brexit
\end{tabular}}
\caption{Sarcasm example from Twitter.}
\label{stab:twitter-example}
\end{subtable}
\vspace{1ex}

\begin{subtable}{\columnwidth}
\centering\resizebox{\columnwidth}{!}{
\begin{tabular}{c||l}
 & \multicolumn{1}{c}{\bf Utterance} \\ 
\hline\hline
\multirow{1}{*}{\texttt{C$_1$}}
  & Promotional images for some guy's Facebook page \\
\hline
\multirow{1}{*}{\texttt{C$_2$}}
 & I wouldn't let that robot near me \\
\hline
\multirow{1}{*}{\texttt{T}}
 & Sounds like you don't like science, you theist sheep\\
\end{tabular}}
\caption{Sarcasm example from Reddit.}
\label{stab:reddit-example}
\end{subtable}

\caption{Examples of the conversation threads where the target utterances involve sarcasm. \texttt{C$_i$}: $i$'th utterance in the context, \texttt{T}: the target utterance.}
\label{tab:sarcasm-examples}
\end{table}

\noindent The Twitter data is collected by using the hashtags \texttt{\#sarcasm} and \texttt{\#sarcastic}.
The Reddit data\LN is a subset of the Self-Annotated Reddit Corpus that consists of 1.3 million sarcastic and non-sarcastic posts \cite{khodak-etal:2017}.
Every target utterance is annotated with one of the two labels, \texttt{SARCASM} and \texttt{NOT\_SARCASM}.
Table~\ref{tab:data-stats} shows the statistics of the two datasets provided by this shared task.

\noindent Notice the huge variances in the utterance lengths for both the Twitter and the Reddit datasets.
For the Reddit dataset, the average lengths of conversations as well as utterances are significantly larger in the test set than the training set that potentially makes the model development more challenging.

\begin{table}[htbp!]
\centering

\begin{subtable}{\columnwidth}
\centering\small 
\begin{tabular}{c||c|c|c}
 & \bf\texttt{NC} & \bf\texttt{AU} & \bf\texttt{AT} \\ 
\hline\hline
\texttt{TRN} & 5,000 & 4.9 ($\pm$3.2) & 140.4 ($\pm$112.8) \\
\texttt{TST} & 1,800 & 4.2 ($\pm$1.9) & 128.5 ($\pm$78.8)$\:\:$ \\
\end{tabular}
\caption{Twitter dataset statistics.}
\label{stab:twitter-stats}
\end{subtable}
\vspace{1ex}

\begin{subtable}{\columnwidth}
\centering\small 
\begin{tabular}{c||c|c|c}
 & \bf\texttt{NC} & \bf\texttt{AU} & \bf\texttt{AT} \\ 
\hline\hline
\texttt{TRN} & 4,400 & 3.5 ($\pm$0.8) & 45.8 ($\pm$17.3) \\
\texttt{TST} & 1,800 & 5.3 ($\pm$2.0) & 93.6 ($\pm$57.8) \\
\end{tabular}
\caption{Reddit dataset statistics.}
\label{stab:reddit-stats}
\end{subtable}

\caption{Statistics of the two datasets provided by the shared task. \texttt{TRN}: training set, \texttt{TST}: test set, \texttt{NC}: \# of conversations, \texttt{AU}: Avg \# of utterances per conversation (including the target utterances) and its stdev, \texttt{AT}: Avg \# of tokens per utterance and its stdev.}
\label{tab:data-stats}
\vspace{-2ex}
\end{table}
\section{Approach}
\label{sec:approach}

Two types of transformer-based sarcasm detection models are used for our experiments:

\begin{enumerate}[label=\alph*)]
\item The target-oriented model takes only the target utterance as input (Section~\ref{ssec:target-oriented}).

\item The context-aware model takes both the target utterance and the context utterances as input (Section~\ref{ssec:context-aware}). 
\end{enumerate}

\noindent These two models are coupled with the latest transformer encoders e.g., BERT \cite{devlin-etal-2019-bert}, RoBERTa \cite{liu-etal-2020-roberta}, and ALBERT \cite{lan-etal-2019-albert}, and compared to evaluate how much impact the context makes to predict whether or not the target utterance involves sarcasm.


\subsection{Target-oriented Model}
\label{ssec:target-oriented}

\begin{figure*}[htbp!]
\centering

\begin{subfigure}{\columnwidth}
\centering
\includegraphics[scale=0.36]{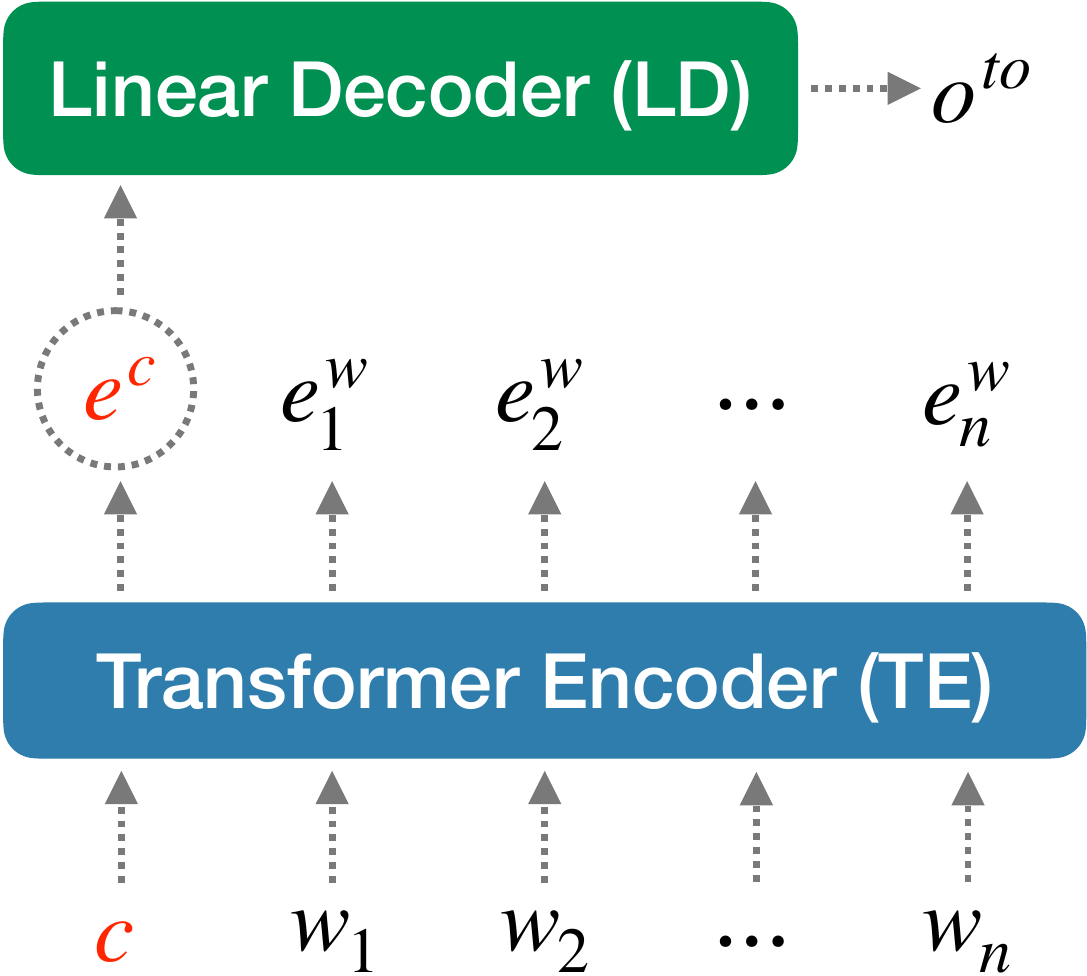}
\caption{Target-oriented model (Section~\ref{ssec:target-oriented})}
\label{sfig:target-oriented-model}
\end{subfigure}
\begin{subfigure}{\columnwidth}
\centering
\includegraphics[scale=0.36]{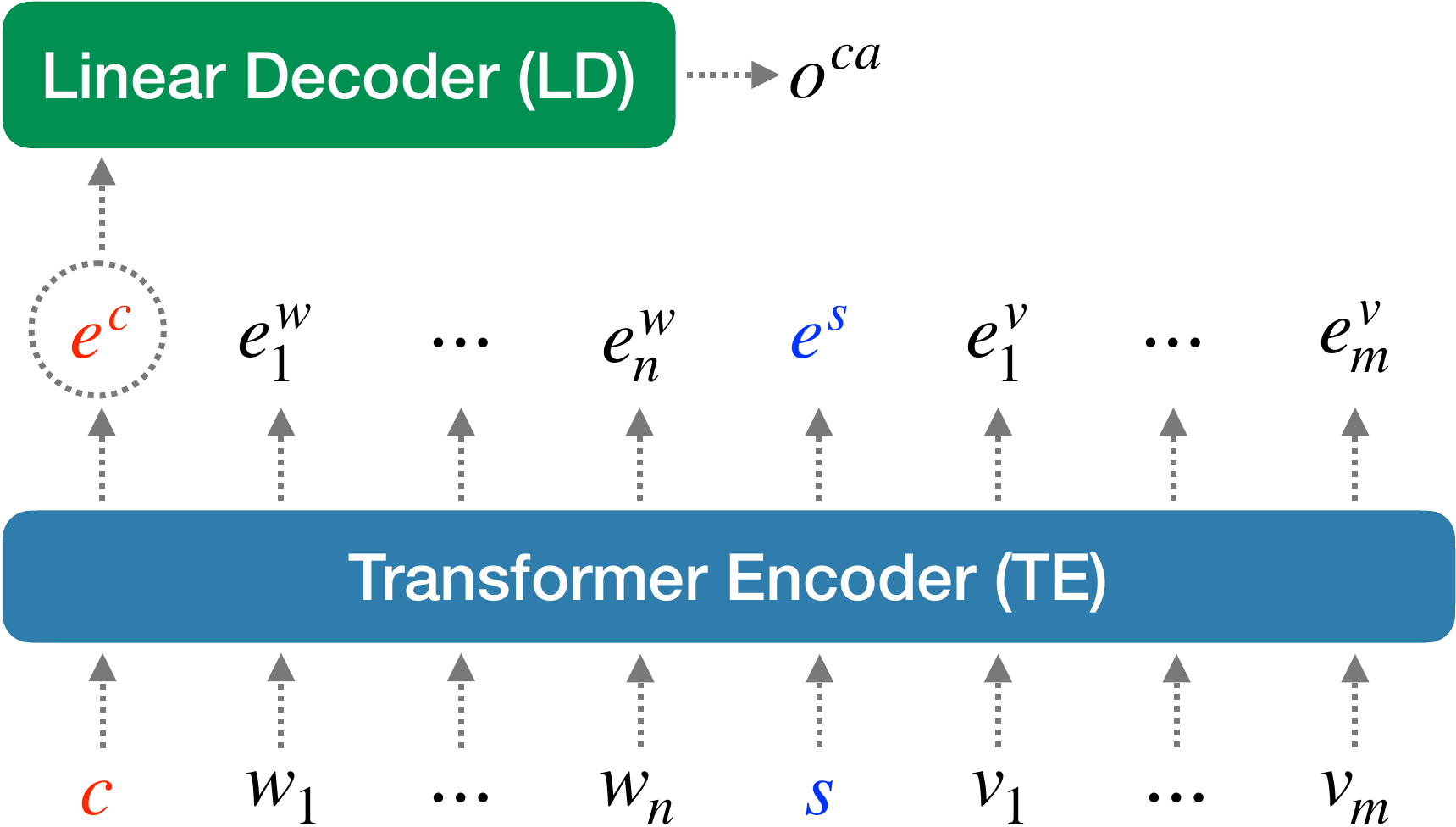}
\caption{Context-aware model (Section~\ref{ssec:context-aware})}
\label{sfig:context-aware-model}
\end{subfigure}

\caption{The overview of our transformer-based target-oriented and context-aware models.}
\label{fig:model-overview}
\end{figure*}

Figure~\ref{sfig:target-oriented-model} shows the overview of the target-oriented model.
Let $W = \{w_{1}, \ldots, w_{n}\}$ be the input target utterance, where $w_{i}$ is the $i$'th token in $W$ and $n$ is the max-number of tokens in any target utterance. 
$W$ is first prepended by the special token $c$ representing the entire target utterance, which creates the input sequence $I^{to} = \{c\} \oplus W$. 
$I^{to}$ is then fed into the transformer encoder, which generates the sequence of embeddings $\{e^c\} \oplus E^w$, where $E^w = \{e^w_{1}, \ldots, e^w_{n}\}$ is the embedding list for $W$ and $(e^c, e^w_{i})$ are the embeddings of $(c, w_{i})$ respectively. 
Finally, $e^c$ is fed into the linear decoder to generate the output vector $o^{to}$ that makes the binary decision of whether or not $W$ involves sarcasm.


\subsection{Context-aware Model}
\label{ssec:context-aware}

Figure~\ref{sfig:context-aware-model} shows the overview of the context-aware model.
Let $L_i$ be the $i$'th utterance in the context.
Then, $V = L_1 \oplus \cdots \oplus L_k = \{v_1, \ldots, v_m\}$ is the concatenated list of tokens in all context utterances, where $k$ is the number of utterances in the context, $v_1$ is the first token in $L_1$ and $v_m$ is the last token in $L_k$.
The input sequence $I^{to}$ from Section~\ref{ssec:target-oriented} is appended by the special token $s$ representing the separator between the target utterance and the context, and also $V$, which creates the input sequence\LN $I^{ca} = I^{to} \oplus \{s\} \oplus V$.
Then, $I^{ca}$ gets fed into the transformer encoder, which generates a sequence of embeddings $\{e^c\} \oplus E^w \oplus \{e^{s}\} \oplus E^{v}$, where $E^{v} = \{e^v_{1}, \ldots, e^v_{m}\}$ is the embedding list for $V$, and $(e^s, e^v_{i})$ are the embeddings of $(s, v_i)$ respectively.
Finally, $e^c$ is fed into the linear decoder to generate the output vector $o^{ca}$ that makes the same binary decision to detect sarcasm.

\section{Experiments}
\label{sec:experiments}

\subsection{Data Split}
\label{ssec:data-split}

For all our experiments, a mixture of the Twitter\LN and the Reddit datasets is used.
The Twitter training set provided by the shared task consists of 5,000 tweets, where the labels are equally balanced between \texttt{SARCASM} and \texttt{NOT\_SARCASM} (Table~\ref{tab:data-stats}).
We find, however, 4.82\% of them are duplicates, which are removed before data splitting.
As a result, 4,759 tweets are used for our experiments.
Labels in the Reddit training set are also equally balanced and no duplicate is found in this dataset.

\vspace{-1ex}
\begin{table}[htbp!]
\centering\small 
\begin{tabular}{c||r|r||r|r}
 & \multicolumn{2}{c||}{\bf Twitter} & \multicolumn{2}{c}{\bf Reddit} \\
\cline{2-5}
 & \multicolumn{1}{c||}{\bf\texttt{TRN}} & \multicolumn{1}{c||}{\bf\texttt{DEV}} & \multicolumn{1}{c||}{\bf\texttt{TRN}} & \multicolumn{1}{c}{\bf\texttt{DEV}} \\ 
\hline\hline
\texttt{SARCASM}      & 2,020 & 239 & 1,973 & 227 \\
\texttt{NOT\_SARCASM} & 2,263 & 237 & 1,987 & 213 \\
\end{tabular}
\caption{Statistics of the data split used for our experiments, where 10\% of each dataset is randomly selected to create the development set.}
\label{tab:train-dev}
\end{table}


\subsection{Models}
\label{ssec:models}

Three types of transformers are used for our experiments, that are BERT-Large \cite{devlin-etal-2019-bert}, RoBERTa-Large \cite{liu-etal-2020-roberta}, and ALBERT-xxLarge \cite{lan-etal-2019-albert}, to compare the performance among the current state-of-the-art encoders.
Every model is run three times and their average scores as well as standard deviations are reported.
All models are trained on the combined Twitter + Reddit training set and evaluated on the combined development set (Table~\ref{tab:train-dev}).


\subsection{Experimental Setup}
\label{ssec:experimental-setup}

After an extensive hyper-parameter search, we set the learning rate to 3e-5, the number of epochs to 30, and use different seed values, 21, 42, 63, for the three runs.
Additionally, based on the statistics of each dataset, we set the maximum sequence length to 128 for the target-oriented models while it is set to 256 for the context-aware models by considering the different lengths of the input sequences required by those approaches. 

\subsection{Results}
\label{ssec:results}

The baseline scores are provided by the organizers, that are 60.0\% for Reddit and 67.0\% for Twitter using the single layer LSTM attention model \cite{ghosh-etal:2018}. 
Table~\ref{tab:results-dev} shows the results achieved by our target-oriented (Section~\ref{ssec:target-oriented}) and the context-aware (Section~\ref{ssec:context-aware}) models on the combined development set.
The RoBERTa-Large model gives the highest F1-scores for both the target-oriented and context-aware models.
The context-aware model using RoBERTa-Large show an improvement of 1.1\% over its counterpart baseline so that this model is used for our final submission to the shared task.
Note that it may be possible to achieve higher performance by fine-tuning hyperparameters for the Twitter and Reddit datasets separately, which we will explore in the future.

\begin{table}[htbp!]
\centering\small

\begin{subtable}{\columnwidth}
\centering
\begin{tabular}{c||c|c|c}
& \bf P & \bf R & \bf F1 \\
\hline\hline
\tt B-L   & 77.3 ($\pm$0.6) & 79.9 ($\pm$0.8) & 78.6 ($\pm$0.1) \\
\tt R-L   & 73.4 ($\pm$0.6) & 88.5 ($\pm$1.4) & \textbf{80.2} ($\pm$0.5) \\
\tt A-XXL & 76.1 ($\pm$1.4) & 83.3 ($\pm$2.3) & 79.5 ($\pm$0.2) \\
\end{tabular}
\caption{Results from the target-oriented models (Section~\ref{ssec:target-oriented}).}
\label{stab:results-dev-to}
\end{subtable}
\vspace{1ex}

\begin{subtable}{\columnwidth}
\centering
\begin{tabular}{c||c|c|c}
& \bf P & \bf R & \bf F1 \\
\hline\hline
\tt B-L   & 76.3 ($\pm$1.0) & 82.7 ($\pm$1.6) & 79.4 ($\pm$0.5) \\
\tt R-L   & 77.3 ($\pm$3.8) & 86.1 ($\pm$4.0) & \textbf{81.3} ($\pm$0.2) \\
\tt A-XXL & 76.5 ($\pm$3.3) & 82.7 ($\pm$3.1) & 79.4 ($\pm$2.2) \\
\end{tabular}
\caption{Results from the context-aware models (Section~\ref{ssec:context-aware}).}
\label{stab:results-dev-ca}
\end{subtable}

\caption{Results on the combined Twitter+Reddit development set. \texttt{B-L}: BERT-Large, \texttt{R-L}: RoBERTa-Large, \texttt{A-XXL}: ALBERT-xxLarge.}
\label{tab:results-dev}
\end{table}

\noindent Table~\ref{tab:results-test} shows the results by the RoBERTa-Large models on the test sets.
The scores are retrieved by submitting the system outputs to the shared task's CodaLab page.\footnote{\url{https://competitions.codalab.org/competitions/22247}}
The context-aware models significantly outperform the target-oriented models on the test sets, showing improvements of 3.1\% and 7.0\% on the F1 scores for the Twitter and the Reddit datasets, respectively.
The improvement on Reddit is particularly substantial due to the much greater lengths of the conversation threads and utterances in the test set compared to the ones in the training set (Table~\ref{tab:data-stats}).
As the final results, we achieve 79.0\% and 75.0\% for the Twitter and Reddit datasets respectively that mark the 2nd places for both datasets at the time of the submission.

\begin{table}[htbp!]
\centering\small

\begin{subtable}{\columnwidth}
\centering
\begin{tabular}{c||c|c|c}
& \bf P & \bf R & \bf F1 \\
\hline\hline
Twitter & 75.5 ($\pm$0.7) & 76.4 ($\pm$0.6) & 75.2 ($\pm$0.8) \\
Reddit  & 67.9 ($\pm$0.5) & 69.2 ($\pm$0.7) & 67.4 ($\pm$0.5) \\
%
\end{tabular}
\caption{Results from the target-oriented RoBERTa-Large models.}
\label{stab:results-test-to}
\end{subtable}
\vspace{1ex}

\begin{subtable}{\columnwidth}
\centering
\begin{tabular}{c||c|c|c}
& \bf P & \bf R & \bf F1 \\
\hline\hline
Twitter & 78.4 ($\pm$0.6) & 78.9 ($\pm$0.3) & \bf{78.3 ($\pm$0.7)} \\
Reddit  & 74.5 ($\pm$0.6) & 74.9 ($\pm$0.5) & \bf{74.4 ($\pm$0.7)} \\
\end{tabular}
\caption{Results from the context-aware RoBERTa-Large models.}
\label{stab:results-test-ca}
\end{subtable}

\caption{Results on the test sets from CodaLab.}
\label{tab:results-test}
\end{table}

\section{Analysis}
\label{sec:analysis}

For a better understanding in our final model, errors from the following three situations are analyzed (\texttt{TO}: target-oriented, \texttt{CA}: context-aware):

\begin{itemize}
\item \texttt{TwCc}: \texttt{TO} is wrong and \texttt{CA} is correct.
\item \texttt{TcCw}: \texttt{TO} is correct and \texttt{CA} is wrong.
\item \texttt{TwCw}: Both \texttt{TO} and \texttt{CA} are wrong.
\end{itemize}


\noindent Table~\ref{tab:prediction-comparision} shows examples for every error situation. 
For \texttt{TwCc}, \texttt{TO} predicts it to be \texttt{NOT\_SARCASM}.
In this example, it is difficult to tell if the target utterance involves sarcasm without having the context.
For \texttt{TcCw}, \texttt{CA} predicts it to be \texttt{NOT\_SARCASM}. 
It appears that the target utterance is long enough to provide enough features for \texttt{TO} to make the correct prediction, whereas considering the extra context may increase noise for \texttt{CA} to make the incorrect decision.
For \texttt{TwCw}, both \texttt{TO} and \texttt{CA} predict it to be \texttt{NOT\_SARCASM}.
This example seems to require deeper reasoning to make the correct prediction.

\begin{table}[htbp!]
\centering
\begin{subtable}{\columnwidth}
\centering\resizebox{\columnwidth}{!}{
\begin{tabular}{c||l}
 & \multicolumn{1}{c}{\bf Utterance} \\ 
\hline\hline
\multirow{1}{*}{\texttt{C$_1$}}
  & who has ever cared about y * utube r * wind . \\
\hline
\multirow{5}{*}{\texttt{C$_2$}}
 & @USER Back when YouTube was beginning it was a \\
 & cool giveback to the community to do a super polished \\
 & high production value video with YT talent . Not the \\
 & same now . The better move for them would be to do like \\
 & 5-6 of them in several categories to give that shine . \\
\hline
\multirow{2}{*}{\texttt{T}}
 & @USER @USER I look forward to the eventual annual \\
 & Tubies Awards livestream .
\end{tabular}}
\caption{Example when \texttt{TO} is wrong and \texttt{CA} is correct.}
\label{stab:c>T example}
\end{subtable}
\vspace{1ex}

\begin{subtable}{\columnwidth}
\centering\resizebox{\columnwidth}{!}{
\begin{tabular}{c||l}
 & \multicolumn{1}{c}{\bf Utterance} \\ 
\hline\hline
\multirow{3}{*}{\texttt{C$_1$}}
  & I am asking the chairs of the House and Senate committees \\
  & to investigate top secret intelligence shared with NBC  \\
  & prior to me seeing it.\\
\hline
\multirow{3}{*}{\texttt{C$_2$}}
 & @USER Good for you, sweetie! But using the legislative\\
 & branch of the US Government to fix your media grudges\\
 & seems a bit much.\\
\hline
\multirow{2}{*}{\texttt{T}}
 & @USER @USER @USER you look triggered after someone \\
 & criticizes me, are conservatives skeptic of ppl in power?\\
\end{tabular}}
\caption{Example when \texttt{TO} is correct and \texttt{CA} is wrong.}
\label{stab:t>c example}
\end{subtable}

\begin{subtable}{\columnwidth}
\centering\resizebox{\columnwidth}{!}{
\begin{tabular}{c||l}
 & \multicolumn{1}{c}{\bf Utterance} \\ 
\hline\hline
\multirow{5}{*}{\texttt{C$_1$}}
  & If I could start my \#Brand over, this is what I would \\
  & emulate my \#Site to look like .. And I might, once my \\
  & anual contract with \#WordPress is up . Even tho I don’t \\
  & think is very; I can’t help but to find ... <URL> <URL>\\
\hline
\multirow{1}{*}{\texttt{C$_2$}}
 & @USER There is no design on it except for links ? \\
\hline
\multirow{4}{*}{\texttt{T}}
 & @USER It's the of what \#Works in this current \#Mindset \\
 & of \#MassConsumption; wannabe fast due to caused by, and \\
 & being just another and. is the light, bringing color back \\
 & to this sad world of and.\\
\end{tabular}}
\caption{Example when both \texttt{TO} and \texttt{CA} are wrong.}
\label{stab:tc example}
\end{subtable}

\caption{Examples of the three error situations. \texttt{C$_i$}: $i$'th utterance in the context, \texttt{T}: the target utterance.}
\label{tab:prediction-comparision}
\vspace{-2ex}
\end{table}

\section{Conclusion}
\label{sec:conclusion}

This paper explores the benefit of considering relevant contexts for the task of sarcasm detection.
Three types of state-of-the-art transformer encoders are adapted to establish the strong baseline for the target-oriented models, which are compared to the context-aware models that show significant improvements for both Twitter and Reddit datasets and become one of the highest performing models in this shared task.

\noindent All our resources are publicly available at Emory NLP's open source repository:
\url{https://github.com/emorynlp/figlang-shared-task-2020}


\section*{Acknowledgments}

We gratefully acknowledge the support of the AWS Machine Learning Research Awards (MLRA).
Any contents in this material are those of the authors and do not necessarily reflect the views of AWS.

\bibliography{acl2020}

\begin{thebibliography}{15}
\expandafter\ifx\csname natexlab\endcsname\relax\def\natexlab#1{#1}\fi

\bibitem[{Bamman and Smith(2015)}]{bamman-smith:2015}
David Bamman and Noah Smith. 2015.
\newblock \href
  {https://www.aaai.org/ocs/index.php/ICWSM/ICWSM15/paper/view/10538}
  {{Contextualized Sarcasm Detection on Twitter}}.
\newblock In \emph{International AAAI Conference on Web and Social Media},
  pages 574--577.

\bibitem[{Devlin et~al.(2019)Devlin, Chang, Lee, and
  Toutanova}]{devlin-etal-2019-bert}
Jacob Devlin, Ming-Wei Chang, Kenton Lee, and Kristina Toutanova. 2019.
\newblock \href {https://www.aclweb.org/anthology/N19-1423} {{BERT:
  Pre-training of Deep Bidirectional Transformers for Language Understanding}}.
\newblock In \emph{Proceedings of the Conference of the North American Chapter
  of the Association for Computational Linguistics: Human Language
  Technologies}, pages 4171--4186.

\bibitem[{Ghosh and Veale(2017)}]{ghosh-veale:2017}
Aniruddha Ghosh and Tony Veale. 2017.
\newblock \href {https://doi.org/10.18653/v1/D17-1050} {{Magnets for Sarcasm:
  Making Sarcasm Detection Timely, Contextual and Very Personal}}.
\newblock In \emph{Proceedings of the 2017 Conference on Empirical Methods in
  Natural Language Processing}, pages 482--491, Copenhagen, Denmark.
  Association for Computational Linguistics.

\bibitem[{Ghosh et~al.(2017)Ghosh, Fabbri, and Muresan}]{ghosh-etal:2017}
Debanjan Ghosh, Alexander~Richard Fabbri, and Smaranda Muresan. 2017.
\newblock \href {http://arxiv.org/abs/1707.06226} {{The Role of Conversation
  Context for Sarcasm Detection in Online Interactions}}.
\newblock \emph{Proceedings of the 18th Annual {SIG}dial Meeting on Discourse
  and Dialogue}, pages 186--196.

\bibitem[{Ghosh et~al.(2018)Ghosh, Fabbri, and Muresan}]{ghosh-etal:2018}
Debanjan Ghosh, Alexander~Richard Fabbri, and Smaranda Muresan. 2018.
\newblock \href {https://doi.org/10.1162/coli_a_00336} {{Sarcasm Analysis using
  Conversation Context}}.
\newblock \emph{Comput. Linguist.}, 44(4):755–792.

\bibitem[{Ghosh et~al.(2015)Ghosh, Guo, and Muresan}]{ghosh-etal:2015}
Debanjan Ghosh, Weiwei Guo, and Smaranda Muresan. 2015.
\newblock \href {https://doi.org/10.18653/v1/D15-1116} {{Sarcastic or Not: Word
  Embeddings to Predict the Literal or Sarcastic Meaning of Words}}.
\newblock In \emph{Proceedings of the 2015 Conference on Empirical Methods in
  Natural Language Processing}, pages 1003--1012, Lisbon, Portugal. Association
  for Computational Linguistics.

\bibitem[{Joshi et~al.(2017)Joshi, Bhattacharyya, and Carman}]{joshi-etal:2017}
Aditya Joshi, Pushpak Bhattacharyya, and Mark~J. Carman. 2017.
\newblock \href {https://doi.org/10.1145/3124420} {{Automatic Sarcasm
  Detection: A Survey}}.
\newblock \emph{ACM Computing Surveys}, 50(5):1--22.

\bibitem[{Joshi et~al.(2015)Joshi, Sharma, and Bhattacharyya}]{joshi-etal:2015}
Aditya Joshi, Vinita Sharma, and Pushpak Bhattacharyya. 2015.
\newblock \href {https://doi.org/10.3115/v1/P15-2124} {{Harnessing Context
  Incongruity for Sarcasm Detection}}.
\newblock In \emph{Proceedings of the 53rd Annual Meeting of the Association
  for Computational Linguistics and the 7th International Joint Conference on
  Natural Language Processing (Volume 2: Short Papers)}, pages 757--762,
  Beijing, China. Association for Computational Linguistics.

\bibitem[{Joshi et~al.(2016)Joshi, Tripathi, Patel, Bhattacharyya, and
  Carman}]{joshi-etal:2016}
Aditya Joshi, Vaibhav Tripathi, Kevin Patel, Pushpak Bhattacharyya, and Mark
  Carman. 2016.
\newblock \href {https://doi.org/10.18653/v1/D16-1104} {{Are Word
  Embedding-based Features Useful for Sarcasm Detection?}}
\newblock In \emph{Proceedings of the 2016 Conference on Empirical Methods in
  Natural Language Processing}, pages 1006--1011, Austin, Texas. Association
  for Computational Linguistics.

\bibitem[{Khodak et~al.(2017)Khodak, Saunshi, and
  Vodrahalli}]{khodak-etal:2017}
Mikhail Khodak, Nikunj Saunshi, and Kiran Vodrahalli. 2017.
\newblock \href {http://arxiv.org/abs/1704.05579} {{A Large Self-Annotated
  Corpus for Sarcasm}}.
\newblock \emph{Proceedings of the Eleventh International Conference on
  Language Resources and Evaluation ({LREC} 2018)}, abs/1704.05579.

\bibitem[{Lan et~al.(2019)Lan, Chen, Goodman, Gimpel, Sharma, and
  Soricut}]{lan-etal-2019-albert}
Zhenzhong Lan, Mingda Chen, Sebastian Goodman, Kevin Gimpel, Piyush Sharma, and
  Radu Soricut. 2019.
\newblock {ALBERT: A Lite BERT for Self-supervised Learning of Language
  Representations}.
\newblock \emph{arXiv}, 11942(1909).

\bibitem[{Liu et~al.(2020)Liu, Ott, Goyal, Du, Joshi, Chen, Levy, Lewis,
  Zettlemoyer, and Stoyanov}]{liu-etal-2020-roberta}
Yinhan Liu, Myle Ott, Naman Goyal, Jingfei Du, Mandar Joshi, Danqi Chen, Omer
  Levy, Mike Lewis, Luke Zettlemoyer, and Veselin Stoyanov. 2020.
\newblock \href {https://openreview.net/forum?id=SyxS0T4tvS} {{RoBERTa: A
  Robustly Optimized BERT Pretraining Approach}}.
\newblock In \emph{Proceedings of the International Conference on Learning
  Representations}.

\bibitem[{Oraby et~al.(2017)Oraby, Harrison, Misra, Riloff, and
  Walker}]{oraby-etal:2017}
Shereen Oraby, Vrindavan Harrison, Amita Misra, Ellen Riloff, and Marilyn
  Walker. 2017.
\newblock \href {https://doi.org/10.18653/v1/W17-5537} {{Are you serious?:
  Rhetorical Questions and Sarcasm in Social Media Dialog}}.
\newblock In \emph{Proceedings of the 18th Annual {SIG}dial Meeting on
  Discourse and Dialogue}, pages 310--319, Saarbr{\"u}cken, Germany.
  Association for Computational Linguistics.

\bibitem[{Van~Hee et~al.(2018)Van~Hee, Lefever, and Hoste}]{hee-etal:2018}
Cynthia Van~Hee, Els Lefever, and V{\'e}ronique Hoste. 2018.
\newblock \href {https://doi.org/10.18653/v1/S18-1005} {{SemEval-2018 Task 3:
  Irony Detection in English Tweets}}.
\newblock In \emph{Proceedings of The 12th International Workshop on Semantic
  Evaluation}, pages 39--50, New Orleans, Louisiana. Association for
  Computational Linguistics.

\bibitem[{Wang et~al.(2015)Wang, Wu, Wang, and Ren}]{Wang-etal:2015}
Zelin Wang, Zhijian Wu, Ruimin Wang, and Yafeng Ren. 2015.
\newblock {Twitter Sarcasm Detection Exploiting a Context-Based Model}.
\newblock In \emph{WISE}.

\end{thebibliography}
\bibliographystyle{acl_natbib}
\end{document}